\pdfoutput=1

\documentclass[11pt]{article}

\usepackage[final]{acl}

\usepackage{times}
\usepackage{latexsym}

\usepackage[T1]{fontenc}

\usepackage[utf8]{inputenc}

\usepackage{microtype}

\usepackage{inconsolata}

\usepackage{graphicx}
\usepackage{multirow}
\usepackage{amsmath}
\usepackage{amssymb}
\usepackage[table,xcdraw]{xcolor}
\usepackage{booktabs}
\usepackage{enumitem}
\usepackage{float}

\usepackage[most]{tcolorbox}

\title{PersonaTrace: Synthesizing Realistic Digital Footprints with LLM Agents}

\author{
 \textbf{Minjia Wang\textsuperscript{1, 2}},
 \textbf{Yunfeng Wang\textsuperscript{1}},
 \textbf{Xiao Ma\textsuperscript{1}},
 \textbf{Dexin Lv\textsuperscript{1}},
 \textbf{Qifan Guo\textsuperscript{1}},
 \textbf{Lynn Zheng\textsuperscript{1}},
\\
 \textbf{Benliang Wang\textsuperscript{1}},
 \textbf{Lei Wang\textsuperscript{1}},
 \textbf{Jiannan Li\textsuperscript{1}},
 \textbf{Yongwei Xing\textsuperscript{1}},
 \textbf{David Xu\textsuperscript{1}},
 \textbf{Zheng Sun\textsuperscript{1}}
\\
\\
 \textsuperscript{1}Apple,
 \textsuperscript{2}Harvard University
\\
 \small{
   \textbf{Correspondence:} \href{mailto:minjiawang@g.harvard.edu}{minjiawang@g.harvard.edu}
 }
}

\begin{document}
\maketitle

\begin{abstract}
Digital footprints—records of individuals’ interactions with digital systems—are essential for studying behavior, developing personalized applications, and training machine learning models. However, research in this area is often hindered by the scarcity of diverse and accessible data. To address this limitation, we propose a novel method for synthesizing realistic digital footprints using large language model (LLM) agents. Starting from a structured user profile, our approach generates diverse and plausible sequences of user events, ultimately producing corresponding digital artifacts such as emails, messages, calendar entries, reminders, etc. Intrinsic evaluation results demonstrate that the generated dataset is more diverse and realistic than existing baselines. Moreover, models fine-tuned on our synthetic data outperform those trained on other synthetic datasets when evaluated on real-world out-of-distribution tasks. 
\end{abstract}

\section{Introduction}

\label{sec:intro}

Digital footprints are the persistent records that individuals leave behind when they interact with digital systems — email threads, chat logs, calendar appointments, purchase histories, sensor traces, and more \citep{shiells2022participant, kolawole2024utilizing}.  

Such traces fuel a wide range of downstream applications: they enable fine-grained user modeling and personalization \citep{valanarasu2021comparative, vullam2023recommendation}, support behavioral and social-science research \citep{golder2014digital-footprints, padricelli2024challenges}, and provide large-scale supervision for data-hungry machine-learning pipelines \citep{zhao2023chatanything}.  

Unfortunately, progress in this area is throttled by data scarcity. 
Publicly available corpora cover only slivers of human activity. For example, the Enron email corpus \citep{klimt2004enron} captures a single company from the early 2000s. They also tend to focus on a single bundle — emails \citep{gholampour2023iwspa, greco2024llmgen}, chat dialogs \citep{zhang2018personachat, suresh2024diasynth, jandaghi2023synthetic-persona-chat}, transaction logs, and so forth — failing to reflect the breadth of modern digital life. Proprietary data are subject to restrictive licenses, as raw digital footprints contain highly sensitive personal information. Regulations such as GDPR prohibit most forms of data sharing, and even internal access is tightly controlled. Anonymization alone is insufficient because rich textual artifacts can be deanonymized with modern LLMs \citep{panda2024steal-pii}.

Synthetic data generation offers a promising workaround and has demonstrated success in training state-of-the-art LLMs \citep{grattafiori2024llama, qwen2025qwen25technicalreport} and addressing tasks such as mathematics \citep{yu2023metamath}, coding \citep{wei2023oss-instruct}, and general instruction following \citep{xu2023wizardlm, li2024aide}. Current synthesis methods, however, presume access to large seed dataset to bootstrap diversity — an assumption that breaks for digital‑footprint data, where both public and private sources are largely inaccessible.

To address these challenges, we introduce \emph{PersonaTrace}, a framework that synthesizes realistic, multi-bundle digital footprints with the help of LLM agents.  
PersonaTrace first creates persona profiles from a pre-defined demographical distribution. 
Given a profile, PersonaTrace simulates a plausible sequence of everyday events (e.g., attending a conference, shopping online, planning a family trip) and then generates the concrete digital artifacts that those events would leave behind (emails, SMS exchanges, calendar entries, reminders, etc).

We assess PersonaTrace with intrinsic metrics that quantify diversity and realism, and with extrinsic metrics that measure downstream utility. Specifically, we fine‑tune open‑source LLMs on the synthetic corpus and evaluate generalization on four real‑world, out‑of‑distribution benchmarks: email categorization, email drafting, question answering, and next‑message prediction. Across tasks, models trained on PersonaTrace achieve competitive or superior results, compared to those trained on the strongest prior synthetic datasets.

Our contributions are as follows:
\vspace{-0.2cm}
\begin{itemize}
    \item We present the first end-to-end method for synthesizing \textit{complete digital footprints} through a persona-driven workflow that ensures coherence and realism across user behaviors.
    \vspace{-0.2cm}
    \item Our comprehensive evaluation demonstrates that our dataset excels in both intrinsic properties such as diversity and realism, and extrinsic performance on downstream tasks.
    \vspace{-0.2cm}
    \item We release \emph{PersonaTrace}, a high-fidelity synthetic digital footprint dataset and accompanying framework, to facilitate responsible future research \footnote{Data and code will be made available on request due to privacy and legal concerns.}.
\end{itemize}


\section{Related Work}

A key strategy for improving quality and diversity for LLM-generated synthetic texts is to guide generation using different priors. 
\vspace{-0.2cm}
\paragraph{Seed–dataset priors.}
Several approaches generate synthetic data by expanding seed datasets, such as conversational corpora \citep{jandaghi2023synthetic-persona-chat}, instruction-tuning training sets \citep{xu2023wizardlm, huang2024datagen, li2024aide, gandhi2024datatune}, or domain-specific problem sets \citep{yu2023metamath, wei2023oss-instruct, braga2024personalized-community-qa, huang2025keypoint, khalil2025creating}. However, these methods are less suited for synthesizing digital footprints due to the lack of comprehensive seed datasets that span multiple modalities (e.g., emails, messages). Publicly available datasets typically focus on a single modality \citep{klimt2004enron, zhang2021emailsum, li-etal-2017-dailydialog, chee2025usticker}, making it difficult to construct coherent multi-modal user profiles.
\vspace{-0.2cm}
\paragraph{LLM-only priors.}
Some approaches rely solely on the generative capabilities of aligned LLMs without using external seed data \citep{xu2024magpie}. While attractive for open-weight checkpoints, commercial APIs typically forbid blank turns or control tokens, limiting its applications.
\vspace{-0.2cm}
\paragraph{Intermediate-attribute priors.}
Some methods guide data generation using intermediate attributes such as knowledge taxonomies or persona descriptions \citep{li2024glan, ge2025scalingsyntheticdatacreation, tang2024matrix-gen, frohling2024attitudes}. However, existing persona-based priors often emphasize professional or academic traits, resulting in biased distributions that do not reflect real-world user profiles and are unsuitable for generating realistic digital footprints.

\section{Methods}
\label{sec:methods}
\begin{figure*}
    \centering
    \includegraphics[width=0.8\linewidth]{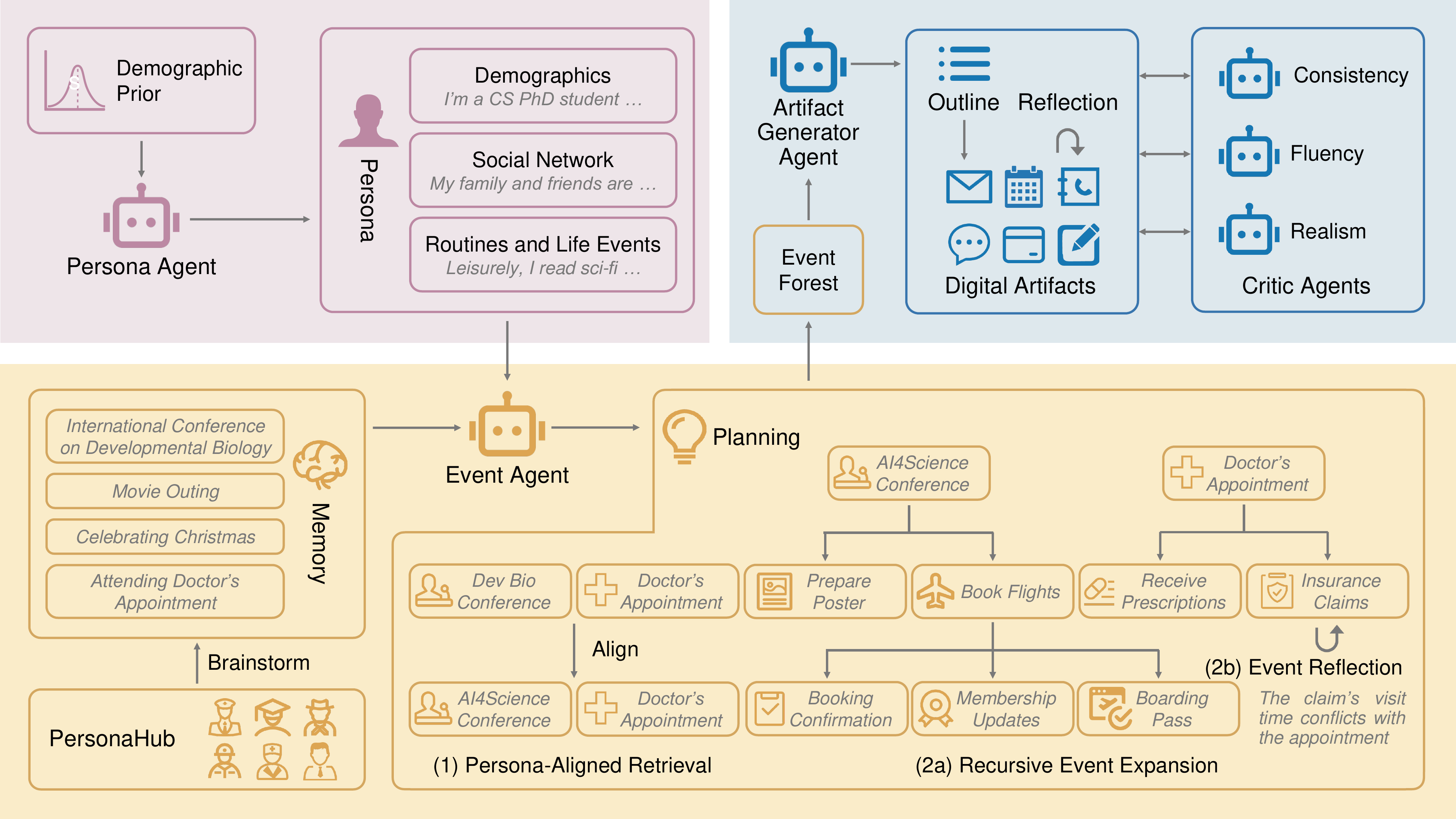}
    \caption{An overview of our methods. The \textbf{Persona Agent} generates a basic profile from a demographic prior, and iteratively adding realistic attributes to it. The \textbf{Event Agent} retrieves seed events from the event memory and aligns them to the persona, and brainstorms with self-reflection to generate an event forest that serves as the scaffolding of the digital footprints. The \textbf{Artifact Generator Agent} and \textbf{Critic Agents} forms a loop - the Artifact Generator Agent generates the outline and then digital artifacts, and the critic agents provides feedback to iteratively improve the quality of the artifacts.}
    \label{fig:framework}
\end{figure*}

\vspace{-0.2cm}
Our approach employs an agent-based architecture built on LLM agents to simulate a realistic user and their digital footprint. We design a three-stage pipeline with specialized autonomous agents that collaborate to generate synthetic data. First, a Persona Agent constructs a detailed persona profile from an initial user specification. Next, an Event Agent expands this persona into a timeline of plausible events tailored to the persona’s life. Finally, an Artifact Generator Agent produces diverse digital artifacts (e.g., emails, text messages, calendar entries, reminders, wallet passes) corresponding to these events, with Critic Agents iteratively reviewing the artifacts for consistency and realism. All agents share the same underlying LLM (Gemini-1.5-Pro with a temperature of 0.9) but are prompted with role-specific instructions and constraints. Figure \ref{fig:framework} shows the overview of our proposed framework. Appendix~\ref{sec:prompts-personatrace} highlights prompts for each agent.

\vspace{-0.05cm}
\subsection{Persona Generation}
\vspace{-0.05cm}
In the first stage, the Persona Agent synthesizes a rich personal profile for the fictional user. We begin by sampling a set of basic demographic attributes from a predefined prior distribution (covering age, gender, locale, etc.) to ground the persona in realism. The priors are estimated from the 2022 American Community Survey \citep{acs2022} to ensure plausible macro-level distributions.  
Given these attributes, the Persona Agent composes a basic identity for the user. This identity includes details such as name, age, gender, birth date, ethnicity, income level, household setup, and location. The Persona Agent then expands this profile by generating a social network context. It creates a list of key relationships (e.g., family members, close friends, coworkers) and assigns each connection basic characteristics (names, ages, relationship to the persona, etc.). To further enhance realism, the Persona Agent outlines the persona’s typical daily routines and major life events. It provides a weekday routine and a weekend routine that reflect the persona’s lifestyle. It also enumerates significant life events such as holidays, vacations, and personal milestones. The outcome of this stage is a comprehensive persona profile that will inform subsequent event generation.

\begin{table*}[!htbp]
\centering
\scriptsize
\begin{tabular}{llccccc}
\toprule
\multicolumn{2}{c}{\textbf{Dataset}} & \textbf{Pairwise Corr. (↓)} & \textbf{Remote-Clique (↑)} & \textbf{Entropy (↑)} & \textbf{Avg. \#Links} & \textbf{Avg. Length} \\
\midrule
 & FinePersonas-Email & 0.2333 & 0.7680 & 2.8080 & 0.0000 & 681.50 \\
 & IWSPA-2023-Adversarial & 0.4079 & 0.5919 & 2.5094 & 0.0000 & 276.48 \\
 & LLM-Gen Phishing & 0.3094 & 0.6906 & 2.6969 & 0.2522 & 685.58 \\
 & Synthetic-Satellite-Emails & 0.5416 & 0.4586 & 2.8218 & 0.0000 & 1050.76 \\
\multirow{-5}{*}{Synthetic} & PersonaTrace (Proposed) & \textbf{0.2093} & \textbf{0.7898} & \textbf{2.8305} & 0.2532 & 1437.87 \\
\midrule
 & Enron & 0.2798 & 0.7218 & 2.7110 & 0.0000 & 2002.57 \\
 & Human-Gen Phishing & 0.1686 & 0.8334 & 2.9257 & 0.0020 & 3332.91 \\
 & Private & 0.2066 & 0.7957 & 2.8332 & 13.6970 & 10036.44 \\
 & Private w/o Spam & 0.2094 & 0.7893 & 2.7079 & 7.6918 & 5814.34 \\
\multirow{-5}{*}{Real} & W3C-Emails & 0.1796 & 0.8196 & 2.7872 & 0.0000 & 2224.89 \\
\bottomrule
\end{tabular}
\caption{Diversity and realism of datasets related to emails. Best results in synthetic datasets are in bold.}
\label{tab:email-stats}
\end{table*}

\begin{table*}[!htbp]
\centering
\scriptsize
\begin{tabular}{llcccccc}
\toprule
\multicolumn{2}{c}{\textbf{Dataset}} & \textbf{Tone} & \textbf{Fluency} & \textbf{Coherence} & \textbf{Informativeness} & \textbf{Engagement} & \textbf{Overall} \\
\midrule
\multirow{5}{*}{Synthetic} & FinePersonas-Email & 4.52 & 4.92 & 4.56 & 3.98 & 3.98 & 4.39 \\
 & IWSPA-2023-Adversarial & 1.59 & 1.91 & 1.67 & 1.31 & 1.22 & 1.53 \\
& LLM-GenPhishing & 3.41 & 4.89 & 4.67 & 3.13 & 2.78 & 3.70 \\
& Synthetic-Satellite-Emails & 4.82 & 4.96 & 4.86 & 4.91 & 3.65 & 4.64 \\
& PersonaTrace (Proposed) &\textbf{ 4.95} & \textbf{4.99} & \textbf{4.99} & \textbf{4.92} & \textbf{4.09} & \textbf{4.79} \\
\midrule
& Enron & 4.19 & 4.73 & 4.47 & 4.28 & 2.71 & 4.07 \\
& Human-Gen Phishing & 3.50 & 4.06 & 3.69 & 3.73 & 2.40 & 3.47 \\
\multirow{-3}{*}{Real} & W3C-Emails & 3.99 & 4.56 & 4.31 & 4.24 & 2.82 & 3.98 \\
\bottomrule
\end{tabular}
\caption{LLM-As-Judge scores of datasets related to emails. Best results in synthetic datasets are in bold.}
\label{tab:email-llm-judge}
\end{table*}

\vspace{-0.05cm}
\subsection{Event Generation}
\vspace{-0.05cm}
\paragraph{Event memory.}
We equip the Event Agent with an internal event memory $\mathcal{M}$ — a knowledge base that stores concise descriptions of activities people experience.  
To populate $\mathcal{M}$, we begin with PersonaHub \citep{ge2025scalingsyntheticdatacreation} as the foundational source, which consists of descriptions of various personas. We then ask the Event Agent to brainstorm plausible daily-life events that a persona in PersonaHub might encounter. The combined list is then pruned for near-duplicates with MinHash LSH \citep{broder1998minhash}, yielding a diverse yet compact collection that serves as the agent’s prior knowledge of the world.

\paragraph{Persona-aligned retrieval.}
Given a persona profile~$\pi$, the Event Agent retrieves a subset of seed events from its memory $\mathcal{M}$: it selects the 30 most semantically relevant entries via embedding search\footnote{Embeddings are obtained from all-MiniLM-L6-v2.}, samples 30 uniformly for diversity, and synthesizes 40 fresh event prompts directly from~$\pi$.  
The Event Agent then aligns each chosen seed event with the persona’s details, modifying event descriptions as needed to fit the persona. 
For example,  
``attend an international conference on developmental biology’’ $\rightarrow$ ``attend an academic conference on AI for science’’ if $\pi$ describes a CS PhD student.

\paragraph{Recursive event expansion.}  
The Event Agent expands each aligned seed event into a tree of sub-events, thereby constructing an event forest $\mathcal{F}=\{\mathcal{T}_1,\dots,\mathcal{T}_n\}$ for the persona. Each seed event serves as a root node in an event tree $\mathcal{T}$, and the Event Agent autonomously decides whether and how to branch that event into finer-grained sub-events. Some events may remain atomic, while others unfold into a sequence of related sub-events. For instance, a travel-related root event like “attend an academic conference on AI for science” might be expanded into sub-events such as “prepare poster” and “book flights”. Then, “book flights” can be further expanded into “receive booking confirmation”, “get membership updates” and “receive boarding pass”. The Event Agent generates these sub-events with appropriate details and temporal ordering, effectively narrating how the larger event plays out. The Event Agent iterates breadth-first until (1) no more sub-events are expanded or (2) $\lvert\mathcal{F}\rvert$ exceeds 300 nodes, whichever comes first.

\vspace{-0.2cm}
\paragraph{Event reflection.} 
Moreover, the Event Agent reflects on each expanded event to ensure quality and completeness. It verifies that the sub-events provide sufficient detail, follow a logical structure, and are likely to result in digital records in the next stage. If any branch is found lacking (e.g., inconsistent), the Event Agent revises the event accordingly. The result of this stage is a collection of event trees – an event forest – that captures a diverse, personalized sequence of events the persona will undergo. This event forest serves as the scaffold for generating digital artifacts in the final stage.

\subsection{Digital Artifact Generation}

In the final stage, the Artifact Generator Agent and the Critic Agents work in tandem to produce high-quality digital artifacts for each event in the event forest. We adopt a generator–critic framework a generator–critic loop à la \citet{madaan2023selfrefine}. For a given event $\varepsilon$ and the persona context $\pi$, the Artifact Generator Agent first creates an outline of the digital artifact $a$ — for example, an email, a text message, a calendar invitation, a reminder, a wallet pass, or other relevant digital record that would reflect what the persona would actually produce or receive in the scenario (e.g., an email confirming a flight booking or a text message exchange with a friend about an upcoming dinner). It then instantiates the artifact based on the outline. Once the Artifact Generator Agent proposes an artifact $a$, each Critic Agent evaluates it and provides feedback. Three Critic Agents evaluate the artifact for consistency with $\varepsilon$ and $\pi$, as well as for its realism and fluency. They ensure that the content aligns with the persona’s known attributes and the details of the event, flagging any contradictions, unnatural language. After receiving the critique, the Artifact Generator Agent revises the artifact $a$ accordingly. This generator–critic loop may repeat multiple times until the artifact meets all quality criteria or a predetermined number of refinement iterations. By the end of this stage, for every event $\varepsilon$ in the persona’s event forest we obtain a finalized digital footprints that documents the event, i.e. $\mathcal{D}_\pi=\{a_1,\dots,a_{|\mathcal{D}|}\}$. Because of the agent-based generation process, the artifacts are not only individually realistic and coherent but also globally consistent with the persona’s life narrative and the sequence of events produced in prior stages.

\section{Evaluation}

\begin{table*}[!htbp]
\centering
\scriptsize
\begin{tabular}{lcccccccccc}
\toprule
\textbf{Training Set} & \multicolumn{2}{c}{\textbf{Enron}} & \multicolumn{2}{c}{\textbf{Human-Gen Phishing}} & \multicolumn{2}{c}{\textbf{Private}} & \multicolumn{2}{c}{\textbf{Private w/o Spam}} & \multicolumn{2}{c}{\textbf{W3C-Emails}} \\

\midrule
\multicolumn{11}{c}{\textbf{Email Categorization}} \\
 & Acc & F1 & Acc & F1 & Acc & F1 & Acc & F1 & Acc & F1 \\
 \midrule
None & 0.2741 & 0.0722 & 0.0818 & 0.0236 & 0.0011 & 0.0022 & 0.0025 & 0.0028 & 0.0011 & 0.0004 \\
FinePersonas-Email & 0.5908 & 0.1810 & \textbf{0.2241} & \textbf{0.0755} & 0.0015 & 0.0027 & 0.0020 & 0.0022 & \textbf{0.5401} & 0.1093 \\
IWSPA-2023-Adversarial & 0.0010 & 0.0038 & 0.0007 & 0.0004 & 0.0012 & 0.0023 & 0.0035 & 0.0039 & 0.0011 & 0.0022 \\
LLM-Gen Phishing & 0.4046 & 0.1095 & 0.1363 & 0.0376 & 0.0008 & 0.0016 & 0.0015 & 0.0017 & 0.1909 & 0.0523 \\
Synthetic-Satellite-Emails & 0.4136 & 0.0848 & 0.1157 & 0.0297 & 0.0009 & 0.0017 & 0.0015 & 0.0017 & 0.2398 & 0.0620 \\
PersonaTrace (Proposed) & \textbf{0.6100} & \textbf{0.1903} & 0.2188 & 0.0659 & \textbf{0.0018} & \textbf{0.0035} & \textbf{0.0051} & \textbf{0.0063} & 0.5311 & \textbf{0.1403} \\

\midrule
\multicolumn{11}{c}{\textbf{Email Drafting}} \\
 & ROUGE & BertS & ROUGE & BertS & ROUGE & BertS & ROUGE & BertS & ROUGE & BertS \\
 \midrule
None & 0.0457 & 0.1175 & 0.0711 & 0.2319 & 0.0545 & 0.1671 & 0.0667 & 0.1818 & 0.1221 & 0.4032 \\
FinePersonas-Email & 0.1541 & 0.4590 & 0.1470 & 0.4835 & 0.1035 & 0.4330 & 0.1246 & \textbf{0.4480} & 0.1704 & \textbf{0.4923} \\
IWSPA-2023-Adversarial & 0.0064 & 0.0170 & 0.0337 & 0.1160 & 0.0072 & 0.0206 & 0.0106 & 0.0281 & 0.0670 & 0.2562 \\
PersonaTrace (Proposed) & \textbf{0.1771} & \textbf{0.4597} & \textbf{0.1599} & \textbf{0.4845} & \textbf{0.1215} & \textbf{0.4337} & \textbf{0.1433} & 0.4429 & \textbf{0.1795} & 0.4744 \\

\midrule
\multicolumn{11}{c}{\textbf{Question Answering}} \\
 & ROUGE & BertS & ROUGE & BertS & ROUGE & BertS & ROUGE & BertS & ROUGE & BertS \\
\midrule 
None & 0.3095 & 0.4766 & 0.2451 & 0.4450 & 0.0425 & 0.3188 & 0.0521 & 0.3265 & 0.1197 & 0.3689 \\
FinePersonas-Email & 0.3203 & 0.4835 & 0.1747 & 0.4207 & 0.0398 & 0.3173 & 0.0451 & 0.3221 & 0.2079 & 0.4263 \\
IWSPA-2023-Adversarial & 0.3904 & 0.5212 & 0.3277 & 0.4984 & 0.0450 & 0.3196 & 0.0535 & 0.3268 & 0.1671 & 0.3956 \\
LLM-Gen Phishing & 0.2333 & 0.4550 & 0.1821 & 0.4447 & 0.0452 & 0.3203 & 0.0547 & 0.3275 & 0.1261 & 0.4086 \\
Synthetic-Satellite-Emails & 0.2540 & 0.4767 & 0.2234 & 0.4734 & 0.0445 & 0.3222 & 0.0529 & 0.3288 & 0.1529 & 0.4288 \\
PersonaTrace (Proposed) & \textbf{0.4435} & \textbf{0.5465} & \textbf{0.4089} & \textbf{0.5313} & \textbf{0.0465} & \textbf{0.3269} & \textbf{0.0559} & \textbf{0.3335} & \textbf{0.2954} & \textbf{0.4643} \\

\bottomrule
\end{tabular}
\caption{Extrinsic evaluation results on email-related downstream tasks: email categorization, email drafting, and question answering. Best results are in bold.}
\label{tab:email-task}
\end{table*}

\vspace{-0.1cm}
\subsection{Baselines}
\vspace{-0.1cm}
We use eight existing synthetic datasets as baselines for comparison. All baseline datasets are described in detail in Appendix~\ref{sec:baselines}.

\vspace{-0.1cm}
\subsection{Intrinsic Evaluation}
\vspace{-0.1cm}

Intrinsic evaluation assesses the inherent properties of the dataset itself.
We use the following metrics to quantitatively measures the diversity of realism of the datasets related to emails.

\textbf{Pairwise Correlation} measures the average Pearson correlation between all pairs of document embeddings. A higher value indicates greater linear correlation, suggesting higher similarity and lower diversity among documents.

\textbf{Remote-Clique} \citep{cox2021diversity} computes the average pairwise cosine distance between document embeddings. Higher values indicate that the documents are more widely dispersed in the embedding space, reflecting greater diversity.

\textbf{Entropy} \citep{cox2021diversity} estimates the Shannon-Wiener entropy of the document embedding distribution. Embeddings are first projected into a 2D space and binned into a $5 \times 5$ grid. The frequency of embeddings in each grid cell is then used to compute entropy. Higher entropy values indicate a more uniform distribution, suggesting greater diversity.

\textbf{Average Number of Links per Email} serves as a proxy for realism, as modern emails typically contain numerous hyperlinks.

\textbf{Average Email Length} is reported for descriptive statistical purposes.

\textbf{LLM-As-Judge Scores} assess human-interpretable and realism-aligned qualities of the emails, including tone, fluency, coherence, informativeness, and engagement. Gemini 2.0 Flash serves as the evaluator, rating each aspect on a 1–5 scale (from poor to excellent). Evaluation prompts are detailed in the Appendix~\ref{sec:llm-as-judge}. Note that, due to privacy constraints, our private datasets were not evaluated, and their corresponding results are therefore omitted.

To improve efficiency, for datasets larger than 1000 samples, we randomly sample a subset of size 1000 for five times, and average metrics across the five samples.

\begin{figure}[H]
\centering
  \includegraphics[width=0.85\columnwidth]{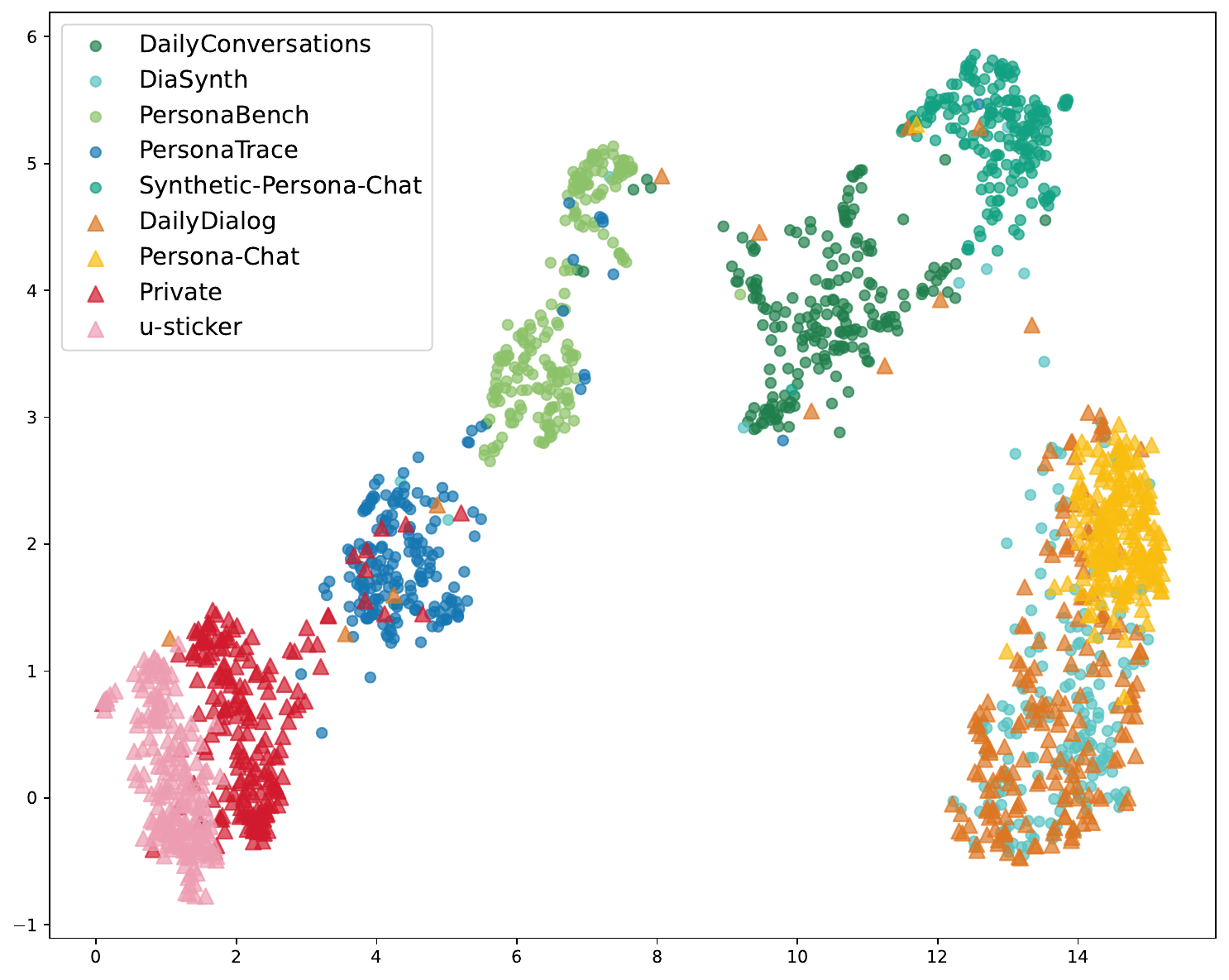}  
  \caption{UMAP visualization of the dataset embeddings related to text messages and conversations. Synthetic datasets are denoted by circles, while real datasets are represented by triangles. Among the synthetic datasets, PersonaTrace appears closest in the embedding space to the private dataset and u-sticker, indicating a higher degree of realism and alignment with real-world digital communications.}
  \label{fig:embedding}
\end{figure}

\begin{table*}[!htbp]
\centering
\scriptsize
\begin{tabular}{lcccccccc}
\toprule
\textbf{Training Set} & \multicolumn{2}{c}{\textbf{DailyDialog}} & \multicolumn{2}{c}{\textbf{Persona-Chat}} & \multicolumn{2}{c}{\textbf{Private}} & \multicolumn{2}{c}{\textbf{u-sticker}} \\
 & Acc & BertS & Acc & BertS & Acc & BertS & Acc & BertS \\
 \midrule
None & \textbf{0.1202} & 0.0237 & 0.1072 & 0.1521 & 0.0958 & 0.1102 & \textbf{0.0933} & 0.0849 \\
DailyConversations & 0.1091 & 0.1867 & 0.1070 & 0.2195 & 0.0961 & 0.2576 & \textbf{0.0933} & 0.2976 \\
DiaSynth & 0.1182 & 0.1334 & 0.1089 & 0.2036 & 0.0961 & 0.1698 & 0.0875 & 0.1666 \\
PersonaBench & 0.1253 & 0.0328 & 0.1064 & 0.1663 & 0.0960 & 0.1246 & 0.0930 & 0.0986 \\
Synthetic-Persona-Chat & 0.1101 & 0.2143 & 0.0982 & 0.2272 & 0.0955 & 0.2954 & 0.0893 & 0.3306 \\
PersonaTrace (Proposed) & \textbf{0.1202} & \textbf{0.2656} & \textbf{0.1150} & \textbf{0.2463} & \textbf{0.0962} & \textbf{0.3178} & 0.0913 & \textbf{0.3526} \\
\bottomrule
\end{tabular}
\caption{Extrinsic evaluation results on message-related downstream task: next message prediction. Best results are in bold.}
\label{tab:messge-task}
\end{table*}

Tables~\ref{tab:email-stats} and~\ref{tab:email-llm-judge} summarize the results. PersonaTrace shows the greatest diversity among synthetic datasets, surpassing even some real ones—such as Enron (across all diversity metrics) and W3C-Emails (in entropy). It also has the longest average email length, reflecting closer resemblance to real data. Furthermore, PersonaTrace attains the highest LLM-as-Judge scores across both synthetic and real datasets, highlighting its superior realism and linguistic quality.

We also provide some straightforward visualization for the generated datasets. For text message related dataset, we use UMAP \citep{mcinnes2018umap} to visualize the embedding.
Figure \ref{fig:embedding} shows the embedding visualization of datasets related to text messages. There are several clusters in the image. The bottom-right cluster reflects daily dialogues and conversation, mainly communicated verbally. The bottom-left cluster, composed of our private dataset and u-sticker, represents the digital communications like text messages or online comments. Our proposed dataset bears close resemblance with the real datasets for digital communication.

\subsection{Extrinsic Evaluation}

\subsubsection{Experiment Setup}

We assess dataset quality by fine-tuning models on synthetic data and evaluating their performance on human-generated benchmarks to measure real-world generalization. Our evaluation focuses on four \textbf{tasks}: email categorization, email drafting, question answering, and next message prediction (see Appendix~\ref{sec:tasks} for task details). While digital footprints extend beyond emails and messages, the lack of high-quality synthetic data in other modalities limits fair comparison.

The \textbf{test datasets} and \textbf{implementation details} are provided in Appendix~\ref{sec:test-datasets} and Appendix~\ref{sec:implementation-details}, respectively.

\subsubsection{Analysis}

 The results are shown in Table \ref{tab:email-task} and \ref{tab:messge-task}.
 
Across all evaluated tasks, PersonaTrace proves to be the most effective synthetic dataset for out-of-distribution generalization. Models fine-tuned on PersonaTrace consistently achieve top performance on both email and dialogue tasks, excelling across metrics such as accuracy, F1, ROUGE-L, and BERTScore. Unlike earlier synthetic datasets that often overfit to narrow domains, PersonaTrace enables models to generalize effectively across varied contexts.

All models struggle on the Private and Private w/o Spam datasets, particularly in email categorization and question answering, where scores remain low. This may be due to the use of a weaker internal model for generating ground-truth labels and the challenging nature of the private emails, which contain many hyperlinks and non-natural language elements (see Table~\ref{tab:email-stats}).

\subsection{Ablation Study}

To evaluate the effectiveness of our agent-based approach, we perform an ablation study by implementing an agent-ablated version of PersonaTrace. In this version, the Event Agent is replaced with a fixed list of predefined event types (e.g., appointments, bills, online shopping, ticketed shows, work meetings), and the LLM is prompted to fill in contextual details such as time and location based on the persona. Additionally, we substitute the Artifact Generator Agent and Critic Agents with a template-based artifact generation process. These templates are manually crafted, with placeholders (e.g., time, location, participants) filled using information from the corresponding event.

\begin{figure}[!h]
\centering
  \includegraphics[width=0.8\columnwidth]{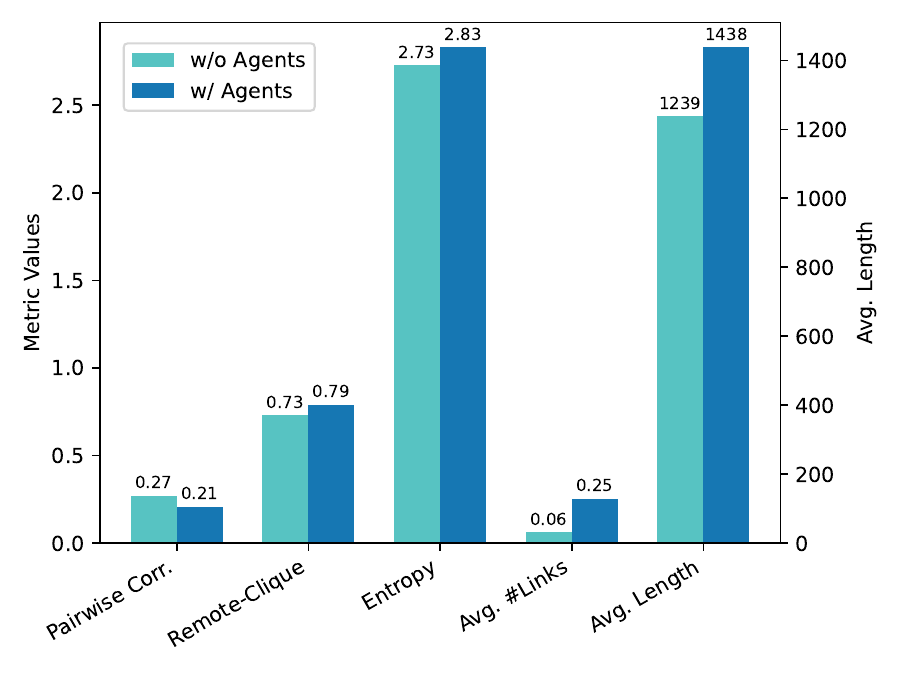}  
  \caption{Comparison of diversity and realism between the agent-ablated and full implementations. For Pairwise Correlation, lower values indicate greater diversity. For Remote-Clique and Entropy, higher values reflect greater diversity. Average number of links per email and average email length are used as indicators of realism, with higher values suggesting closer resemblance to human-generated emails.}
  \label{fig:ablation}
\end{figure}

Figure~\ref{fig:ablation} and Table~\ref{tab:ablation} present the results of the ablation study. The full agent-based implementation outperforms the template-based baseline in both diversity and realism. It also achieves significantly better performance on downstream tasks, demonstrating superior generalizability.

\begin{table}[H]
\scriptsize
\centering
\begin{tabular}{c ccc}
\toprule
\multicolumn{2}{c}{\textbf{Task}} & \textbf{w/o Agents} & \textbf{w/ Agents} \\
\midrule
 & Acc & 0.0063 & 0.2733 \\
\multirow{-2}{*}{Email Categorization} & F1 & 0.0057 & 0.0813 \\
 & ROUGE & 0.0376 & 0.1527 \\
\multirow{-2}{*}{Email Drafting} & BertS & 0.3012 & 0.4590 \\
 & ROUGE & 0.0413 & 0.2500 \\
\multirow{-2}{*}{Question Answering}& BertS & 0.2880 & 0.4405 \\
 & Acc & 0.1015 & 0.1056 \\
\multirow{-2}{*}{Next Utterance Prediction} & BertS & 0.1938 & 0.2956 \\
\bottomrule
\end{tabular}
\caption{Comparison of downstream task performance between the agent-ablated and full implementations. }
\label{tab:ablation}
\end{table}

\section{Conclusion}

We introduced PersonaTrace, the first end-to-end pipeline for generating realistic synthetic digital footprints using LLM agents. Starting from high-level persona profiles, our framework produces diverse and plausible user events along with their corresponding digital artifacts, including emails, chat messages, calendar entries, wallet passes, reminders, etc. PersonaTrace offers high diversity and realism, making it well-suited for training models on a wide range of downstream tasks. We train our model on the dataset and deploy it in our online retrieval product, achieving a 7\% absolute improvement in Recall@10. The dataset and generation framework will be released after passing internal review to benefit the community.


\section*{Limitations}

\textbf{Limited control over artifact topics.} The current implementation relies heavily on the Event Agent, which constructs event trees based on the LLM's prior knowledge. As a result, it is difficult to constrain or guide the generation toward specific topics (for instance, specifying that all artifacts should relate to travel). Future work can focus on enhancing controllability over artifact content (e.g., topic or intent) for specific applications.

\section*{Ethical Considerations}

\paragraph{Privacy and data protection.}
PersonaTrace has been designed with ethical safeguards at its core. Importantly, the framework relies exclusively on synthetic data generated by agentic LLMs, ensuring that no real user information is collected or exposed. The project has undergone and passed our institution’s internal privacy and legal compliance review, overseen by the internal legal and privacy team.

\paragraph{Responsible use and access control.}
While analyses of digital footprints can offer significant benefits for user experience, digital assistant effectiveness research and behavioral research, they can also be misapplied to surveillance or the prediction of protected attributes. To mitigate these risks, we will release both the dataset and framework under terms and licenses that explicitly prohibit applications related to surveillance or inferring protected groups. In addition, access to the dataset and codebase will require users to agree to these conditions, thereby aligning usage with principles of responsible research and beneficial impact.

\paragraph{Bias awareness and fair representation.}
PersonaTrace estimates population priors from the 2022 American Community Survey, anchoring the generation process in empirically grounded and demographically representative distributions. This prevents the emergence of arbitrary or systematically skewed data.
In addition, our framework supports the dynamic inclusion of specific demographic backgrounds through manually crafted personas, allowing careful control over representation. 

By incorporating these safeguards, we aim to maximize the positive research potential of PersonaTrace while reducing the risk of harmful or unethical applications.

\bibliography{custom}

\appendix

\label{sec:appendix}

\section{Baselines}
\label{sec:baselines}

\subsection{Synthetic Emails}

\textbf{FinePersonas-Email}\footnote{https://huggingface.co/datasets/argilla/FinePersonas-Synthetic-Email-Conversations} comprises approximately 114,000 synthetic email exchanges between pairs of personas. It is created by selecting around 11,000 personas from FinePersonas-v0.1. Each is paired with five semantically similar and five random personas to generate diverse scenarios. The Hermes-3-Llama-3.1-70B model then uses chain-of-thought reasoning to produce context-rich email exchanges for each pair. 

\textbf{IWSPA-2023-Adversarial} \citep{gholampour2023iwspa} contains 5,000 synthetic adversarial emails. This dataset was created by applying four adversarial text attack techniques—TextFooler, PWWS, DeepWordBug, and BAE—from the TextAttack framework to the IWSPA 2.0 dataset.

\textbf{LLM-Gen Phishing} \citep{greco2024llmgen} consists of 1,000 legitimate emails generated by ChatGPT, and 1,000 emails generated by WormGPT.

\textbf{Synthetic-Satellite-Emails}\footnote{https://huggingface.co/datasets/KeystoneIntelligence/\newline synthetic-satellite-conjunction-emails} contains 1,200 synthetic email communications related to satellite conjunction scenarios.

\subsection{Synthetic Messages and Conversations}

\textbf{DailyConversations}\footnote{https://huggingface.co/datasets/safetyllm/dailyconversations} is synthetically generated using ChatGPT 3.5 and comprises two-person multi-turn dialogues covering various topics. It has nearly 31,000 dialogues in total.

\textbf{DiaSynth} \citep{suresh2024diasynth} is a synthetic dialogue generation framework tailored for low-resource applications, using LLMs including Phi-3, InternLM-2.5, LLaMA-3 and GPT-4o and Chain-of-Thought reasoning to produce persona-driven dialogues. The dataset contains 13,000 dialogues.

\textbf{Synthetic-Persona-Chat} \citep{jandaghi2023synthetic-persona-chat} is a persona-based conversational dataset, extending the original Persona-Chat dataset with new synthetic conversations. It contains 21,907 conversations. This dataset is generated using a Generator-Critic framework to ensure the quality and faithfulness of the dialogues.

\textbf{PersonaBench} \citep{tan2025personabench} involves a synthetic data generation pipeline that creates diverse, realistic user profiles and private conversations simulating human activities. It contains 1,600 conversations generated by GPT-4o.

\section{Prompts of LLM-As-Judge}
\label{sec:llm-as-judge}

Please refer to Figure~\ref{fig:llm-as-judge-2}.

\begin{figure*}[ht]
\centering
\scriptsize
\begin{tcolorbox}[colback=blue!5!white,colframe=blue!75!black,title=LLM-As-Judge]
You are an expert evaluator for synthetic communication data.

Your task is to evaluate the following email based on multiple quality dimensions.

Carefully read the email content and provide structured ratings and feedback.

\vspace{0.5cm}
\textbf{Evaluation Dimensions}

\begin{enumerate}
  \item Tone
  \begin{itemize}
    \item Is the tone appropriate for the context?
    \item Is it consistent throughout the email?
    \item Is it aligned with the intended audience?
  \end{itemize}
  \item Fluency
  \begin{itemize}
    \item Is the writing smooth and grammatically correct?
    \item Does it sound natural to read?
  \end{itemize}
  \item Coherence
  \begin{itemize}
    \item Are the ideas logically connected?
    \item Is the email easy to follow?
  \end{itemize}
  \item Informativeness
  \begin{itemize}
    \item Does the text provide useful, accurate, and complete information?
    \item Does it avoid missing or misleading details?
  \end{itemize}
  \item Engagement
  \begin{itemize}
    \item Does the text capture and maintain the reader’s attention?
    \item Does it encourage the reader to take action if needed?
  \end{itemize}
\end{enumerate}

\textbf{Scoring Guideline (for each dimension)}

\begin{itemize}[label={}]
    \item 5 = Excellent: Fully meets requirements, no issues.
    \item 4 = Good: Mostly meets requirements, with minor flaws.
    \item 3 = Fair: Some issues present, partially acceptable.
    \item 2 = Poor: Major issues, mostly unacceptable.
    \item 1 = Very Poor: Completely fails the requirement, unusable.
\end{itemize}

\textbf{Output Requirements}

Give a 1–5 score for each dimension with a short explanation (1–2 sentences).
Provide an overall evaluation with an overall score (average or holistic).
Use JSON format for the output.

\textbf{Example Output}

\{

\quad"Tone": \{

\qquad "score": 4,

\qquad "explanation": "Tone is polite and suitable for a business email, but slightly too formal for the intended young audience."

\quad \},

\quad "Fluency": \{

\qquad "score": 5,

\qquad "explanation": "Grammar and flow are flawless; very natural phrasing."

\quad \},

\quad "Coherence": \{

\qquad "score": 4,

\qquad "explanation": "Message is generally easy to follow, though one sentence feels abrupt."

\quad\},

\quad "Informativeness": \{

\qquad "score": 5,

"explanation": "All key details are included and accurate."

\quad\},

\quad "Engagement": \{

\qquad "score": 3,

\qquad "explanation": "The message provides information but lacks a strong hook to engage the reader."

\quad \},

\quad "Overall": \{

\qquad "score": 4.2,

\qquad "summary": "Well-written and informative, but slightly formal and could be more engaging."

\quad \}

\}

\textbf{Input}

input: \{input\}
\end{tcolorbox}
\caption{Prompts for LLM-As-judge evaluation.}
\label{fig:llm-as-judge-2}
\end{figure*}

\section{Prompts of PersonaTrace}
\label{sec:prompts-personatrace}

In this section, we present several representative prompts used in the PersonaTrace framework to enhance transparency and facilitate reproducibility.

\paragraph{Persona Agent.}
Figures~\ref{fig:protagonist-concept-1} and~\ref{fig:protagonist-concept-2} illustrate the prompt used by the Persona Agent to enrich basic demographic information with detailed and realistic personal attributes. The agent refines the sampled demographic features, which are drawn from a real-world, statistics-grounded distribution, into coherent, lifelike persona profiles.

\paragraph{Event Agent.}
The Event Agent is responsible for constructing an event tree by expanding a set of seed events. Seed events can be obtained in three ways: (1) uniform sampling from the event memory, (2) retrieving the most similar events from the event memory, or (3) directly generating events from the persona profile. Figure~\ref{fig:event-brainstorming} presents the prompt for the third approach. During event expansion, the prompt shown in Figure~\ref{fig:event-expansion} guides the iterative process of developing each event into multiple sub-events, ensuring temporal and causal consistency.

\paragraph{Artifact Generator Agent.}
We use email generation as an illustrative example of the Artifact Generator Agent. The process begins with producing an outline of the email (Figure~\ref{fig:outline-generation}), followed by generating the full email content based on both the outline and the associated event (Figure~\ref{fig:email-generator}). The reflection phase involves two stages: first, the model generates constructive feedback on the initial draft (Figure~\ref{fig:email-review}); then, it revises the email according to this feedback (Figure~\ref{fig:email-revision}).
\begin{figure*}[ht]
\centering
\scriptsize
\begin{tcolorbox}[colback=blue!5!white,colframe=blue!75!black,title=Profile Generation]
\textbf{Role:}  
You are tasked with writing a novel that captures life in the modern world.

\textbf{Mission:}  
Your primary task is to develop a detailed concept for your novel's protagonist. This includes articulating specifics about their job, personal life, and social connections. You must organize and present this concept in a JSON format.

\textbf{Task Requirements:}
\begin{enumerate}
    \item Populate each provided field relevant to the protagonist's life, including personal characteristics and daily routines. If a specific field (e.g., \texttt{classmates}) does not apply to your character design, omit this field entirely.
    \item Any information you include must align with the initial input. If additional information is necessary and was not provided in the input, extrapolate reasonably based on the available data. Avoid using placeholders such as “not specified” or seeking further clarification.
    \item Choose a name for your protagonist reflecting their gender and ethnicity to ensure authenticity and sensitivity.
    \item Factor in the protagonist's income level when outlining their lifestyle, specifically their holiday and vacation activities.
    \item Ensure all content is original and, when formatting your response, reference only the structure—not the content—of provided examples.
    \item All output keys should be in English, and all values should be in the user's local language.
    \item The protagonist's nationality should reflect only the nationality indicated on their passport, while the protagonist's residence address must correspond to the specified \texttt{locale} in the input.
\end{enumerate}

\textbf{Input:}  
The protagonist's profile should be JSON formatted and include:
\begin{itemize}
    \item \texttt{name}: the protagonist's full name
    \item \texttt{locale}: language and geographic location
    \item \texttt{timezone}: local timezone
    \item \texttt{age}: age of the protagonist (string value)
    \item \texttt{gender}: gender identity
    \item \texttt{income}: income bracket
    \item \texttt{ethnicity}: ethnic background
    \item \texttt{family\_setup}: description of familial relationships
    \item \texttt{nationality}: the protagonist's nationality
\end{itemize}

\textbf{Output:}  
Your output should be a detailed JSON formatted document expanding upon the input and including additional fields such as:
\begin{itemize}
    \item \texttt{surname}: protagonist’s surname, resolved from the full name.
    \item \texttt{given\_name}: protagonist’s given name, resolved from the full name.
    \item \texttt{middle\_name}: protagonist’s middle name (if any), resolved from the full name. Omit this field if inapplicable.
    \item \texttt{nicknames}: list of protagonist’s nicknames, in the user’s local language.
    \item \texttt{email}: randomly generated email address using realistic username and domain conventions based on locale.
    \item \texttt{phone}: random generated phone number adhering to the locale’s format.
    \item \texttt{eye\_color}: one of [black, blue, brown, gold, gray, green, silver, white].
    \item \texttt{hair\_color}: one of [black, blue, brown, gold, gray, green, silver, white].
    \item \texttt{height}: physical height.
    \item \texttt{weight}: physical weight.
    \item \texttt{occupation}: detailed job role, written in the user’s local language.
    \item \texttt{weekdays\_routines}: narrative of a typical weekday, written in the local language.
    \item \texttt{weekend\_routines}: narrative of a typical weekend, written in the local language.
    \item \texttt{life\_events\_for\_holidays\_and\_vacations}: description of holidays and vacation practices, written in the local language.
\end{itemize}
\end{tcolorbox}
\caption{Prompt for generating comprehensive and culturally grounded profile.}
\label{fig:protagonist-concept-1}
\end{figure*}

\begin{figure*}[ht]
\centering
\scriptsize
\begin{tcolorbox}[colback=blue!5!white,colframe=blue!75!black,title=Profile Generation (Continued)]
\begin{itemize}
    \item \texttt{family\_members}: list including names, ages, relations, occupations, and workplace/school addresses—all in the local language, with realistic naming conventions for the locale.
    \item \texttt{friends}: list of five friends’ names in the local language, with culturally correct name order and spacing.
    \item \texttt{coworkers}: list of eight coworkers’ names in the local language, with proper format.
    \item \texttt{classmates}: if applicable, list of ten classmates’ names in the local language, formatted correctly.
    \item \texttt{home\_address}: realistic residential address in the local language, aligned with the locale.
    \item \texttt{office\_address}: realistic office address in the local language, aligned with the locale (omit if inapplicable).
    \item \texttt{school\_address}: realistic school address in the local language (omit if inapplicable).
\end{itemize}
\end{tcolorbox}
\caption{Prompt for generating comprehensive and culturally grounded profile.}
\label{fig:protagonist-concept-2}
\end{figure*}

\begin{figure*}[ht]
\centering
\scriptsize
\begin{tcolorbox}[colback=blue!5!white,colframe=blue!75!black,title=Seed Events Generation]
\textbf{Task}  
Brainstorm possible events based on the profile. Consider all possibilities, and generate at least \{\texttt{num\_seed\_events}\} events as comprehensive and diverse as possible. 

Here are some tips for brainstorming:
\begin{itemize}
    \item \textbf{Analyze Lifestyle.} Identify daily, weekly, and seasonal patterns. Consider work, hobbies, social life, and personal responsibilities.
    \item \textbf{Consider Recent Life.} Reflect on important events in the past two years.
    \item \textbf{Incorporate Professional and Personal Roles.} Include work-related tasks. Consider personal interests.
    \item \textbf{Account for Special Occasions and Holidays.} Include holiday traditions, family gatherings, and vacations. Consider birthdays, anniversaries, and cultural events.
    \item \textbf{Think About Common Responsibilities.} Cover financial management. Include household chores.
    \item \textbf{Consider Social and Recreational Activities.} Identify interactions with family, friends, and coworkers. Include leisure activities like travel, hobbies, or fitness.
    \item \textbf{Factor in Unexpected and Rare Events.} Account for emergencies (e.g., medical visits, car repairs). Consider special projects or one-time commitments.
\end{itemize}

\vspace{0.5cm}
\textbf{Output Format}  

A JSON object with the following fields:
\begin{itemize}
    \item \texttt{event}: A clear and specific event title.
    \item \texttt{detailed\_description}: A comprehensive explanation of the event for consistency and coherence.
    \item \texttt{frequency}: A string representing how often the event occurs, chosen from the predefined options: [``daily'', ``weekly'', ``monthly'', ``seasonally'', ``yearly'', ``once''].
\end{itemize}

\vspace{0.5cm}
\textbf{Input}  
\{\texttt{profile}\}

\vspace{0.3cm}
\textbf{Output}  
Let's think step by step. \\
First, I need to break down the weekday and weekend routines into a list of events. Second, I need to brainstorm for events in the recent life.
\end{tcolorbox}
\caption{Prompt for brainstorming comprehensive, profile-based events.}
\label{fig:event-brainstorming}
\end{figure*}

\begin{figure*}[ht]
\centering
\scriptsize
\begin{tcolorbox}[colback=blue!5!white,colframe=blue!75!black,title=Event Expansion]
\textbf{Input Format:}  

You will receive a JSON object, representing an event with the following fields:  
\begin{itemize}
    \item \texttt{event}: A clear and specific event title.  
    \item \texttt{detailed\_description}: A comprehensive explanation of the event to ensure consistency and coherence.  
    \item \texttt{frequency}: How often the event occurs—one of: [``daily'', ``weekly'', ``monthly'', ``seasonally'', ``yearly'', ``once''].
    \item \texttt{location}: A realistic and precise address that fits the event, suitable for a calendar entry. If the event could take place in multiple locations, it is left blank.  
    \item \texttt{other\_participants}: A list of attendees, selected only from the names provided in the profile. If no additional participants are needed, it is left blank.  
    \item \texttt{start\_time}: The start time in RFC3339 format without a time zone.  
    \item \texttt{end\_time}: The end time in RFC3339 format without a time zone.  
\end{itemize}

\vspace{0.5cm}
\textbf{Your Task:}  

You must analyze the event and brainstorm relevant events as comprehensively as possible. Here are some tips:
\begin{itemize}
    \item \textbf{Think of All Possible Variations.} Account for different circumstances. Consider different methods or approaches. Consider various subcategories.
    \item \textbf{Consider Different Perspectives.} Look at the event from a personal, professional, logistical, and financial angle.
    \item \textbf{Include Decision Points and Contingencies.} Consider what happens if something goes wrong. Identify common problems and possible solutions.
    \item \textbf{Cover Tools, Resources, and External Interactions.} Mention necessary tools. Identify people involved.
\end{itemize}

\vspace{0.5cm}
\textbf{Output Format:}  

A list of JSON objects, representing relevant events with the following fields:  
\begin{itemize}
    \item \texttt{event}: A clear and specific event title.  
    \item \texttt{detailed\_description}: A comprehensive explanation of the event to ensure consistency and coherence.  
    \item \texttt{frequency}: How often the event occurs—one of: [``daily'', ``weekly'', ``monthly'', ``seasonally'', ``yearly'', ``once''].
    \item \texttt{location}: A realistic and precise address that fits the event, suitable for a calendar entry. If the event could take place in multiple locations, leave this blank. You can reference locations from the profile or suggest reasonable alternatives.
    \item \texttt{other\_participants}: A list of attendees, selected only from the names provided in the profile. If no additional participants are needed, leave this blank.
    \item \texttt{start\_time}: The start time in RFC3339 format without a time zone.  
    \item \texttt{end\_time}: The end time in RFC3339 format without a time zone.  
\end{itemize}

\vspace{0.5cm}
\textbf{Examples:}  
\{examples\}

\vspace{0.5cm}
\textbf{Your Turn}  

\textbf{Input:}  
\{input\_event\}

\vspace{0.3cm}
\textbf{Output:}
\end{tcolorbox}
\caption{Prompt for generating related event expansions from a single event description.}
\label{fig:event-expansion}
\end{figure*}

\begin{figure*}[ht]
\centering
\scriptsize
\begin{tcolorbox}[colback=blue!5!white,colframe=blue!75!black,title=Email Outline Generation]
\textbf{Task:}  

You are a specialist in creating emails. You will be provided with a JSON object representing an event. 
Your objective is to generate a realistic outline for the \textbf{body} of the email that \{full\_name\} \textbf{\{sent\_or\_received\}}.

\vspace{0.5cm}
\textbf{Event Details (JSON):}  
\{event\}

\textbf{Note:} Some fields are guaranteed to be present (\texttt{event}, \texttt{detailed\_description}, \texttt{start\_time}, \texttt{end\_time}, \texttt{location}, \texttt{other\_participants}), while others are optional and should only be used if relevant.

\vspace{0.5cm}
\textbf{Instructions:}  
\begin{enumerate}
    \item Output a \textbf{detailed outline} (not a fully written email) of the \textbf{sender's} email.  
    \item You do not need to use all JSON fields, just those that make sense for the context of the email.
    \item Highlight any actions, requests, or follow-up details needed from the recipients.  
    \item Choose an appropriate tone suitable for the event context.
    \item Do not include placeholder text. Instead, use actual data or reasonable, context-based values.
    \item You may include additional resources or references, if applicable.
\end{enumerate}

\vspace{0.5cm}
\textbf{Final Deliverable:}  
\begin{itemize}[label={}]
    \item Provide a structured outline (like headings and bullet points) of the email body that \{full\_name\} \{sent\_or\_received\}.
    \item The outline should reflect the \textbf{sender's} viewpoint.
\end{itemize}

\vspace{0.5cm}
\textbf{Outline:}
\end{tcolorbox}
\caption{Prompt for generating structured email body outlines based on event data.}
\label{fig:outline-generation}
\end{figure*}

\begin{figure*}[ht]
\centering
\scriptsize
\begin{tcolorbox}[colback=blue!5!white,colframe=blue!75!black,title=Email Gneration]
You are a specialist in writing emails. You will be provided with an outline of an email along with additional reference content. Your job is to craft a realistic, engaging, and well-structured email based on the outline.

\vspace{0.5cm}
\textbf{Instructions:}

\begin{enumerate}
    \item \textbf{Input Details:}
    \begin{itemize}
        \item \textbf{Outline:} You will receive an outline of the email, which includes the main points and structure to cover.
        \item \textbf{Additional Reference:} You will also be provided with a JSON object containing event-related details. The fields that are always present are:
        \begin{itemize}
            \item \texttt{event}
            \item \texttt{detailed\_description}
            \item \texttt{start\_time}
            \item \texttt{end\_time}
            \item \texttt{location}
            \item \texttt{other\_participants}
        \end{itemize}
        \item Other fields in the JSON object are optional. \textbf{Note:} Use only the relevant fields to create a clear and effective email.
        \item \textbf{Note:} You do not need to incorporate every field from the JSON object; only use the information that is relevant to create a clear and effective email.
    \end{itemize}

    \item \textbf{Email Composition Guidelines:}
    \begin{itemize}
        \item \textbf{Structure \& Tone:}
        \begin{itemize}
            \item Write a realistic and engaging email that follows the provided outline.
            \item Choose a tone that matches the context of the event and the intended recipients. For example, for an emergency preparedness notice, use a calm, reassuring, and informative tone; for a celebratory event, a more upbeat tone is suitable.
        \end{itemize}
        \item \textbf{Content Integration:}
        \begin{itemize}
            \item Use the outline as the framework for your email body.
            \item Incorporate relevant details from the additional reference JSON object to enhance the content.
            \item Ensure the email includes critical event information such as event name, detailed description, dates, location, and any important context provided.
        \end{itemize}
        \item \textbf{Clarity and Readability:}
        \begin{itemize}
            \item Organize the email into clear sections based on the outline.
            \item Use headings, paragraphs, and bullet points where appropriate to enhance readability.
        \end{itemize}
        \item \textbf{Relevance:}
        \begin{itemize}
            \item Only include information from the JSON object that directly contributes to the purpose and clarity of the email.
            \item Avoid unnecessary details that do not add value or could distract from the main message.
            \item You may add extra details that complement the outline and reference material if needed.
        \end{itemize}
        \item \textbf{Call-to-Action:}
        \begin{itemize}
            \item Including specific next steps or call-to-action is optional. Only include them if they enhance the clarity and usefulness of the email.
        \end{itemize}
    \end{itemize}

    \item \textbf{Output Structure:}
    \begin{itemize}
        \item The final email must be structured as a JSON object with the following keys:
        \begin{itemize}
            \item \texttt{sender\_name}: The name of the sender.
            \item \texttt{from\_address}: The sender's email address.
            \item \texttt{to\_address}: The receiver's email address.
            \item \texttt{send\_time}: The time the email is sent in RFC3339 format without a time zone.
            \item \texttt{subject}: A concise and relevant subject line.
            \item \texttt{body}: The complete email body text, following the outline.
        \end{itemize}
    \end{itemize}

    \item \textbf{Process:}
    \begin{itemize}
        \item Start by reviewing the provided outline and event reference.
        \item Develop a cohesive email that aligns with the outline and appropriately integrates relevant event details.
        \item Ensure that the email is organized, clear, and engaging, following standard email conventions.
    \end{itemize}
\end{enumerate}

\vspace{0.5cm}
\textbf{Outline:}  
\{outline\}

\vspace{0.3cm}
\textbf{Additional References:}  
\{event\}
\end{tcolorbox}
\caption{Prompt for generating realistic emails from outlines and event references.}
\label{fig:email-generator}
\end{figure*}

\begin{figure*}[ht]
\centering
\scriptsize
\begin{tcolorbox}[colback=blue!5!white,colframe=blue!75!black,title=Email Review]
You are an expert in email review and writing. I will provide you with an email, and I need you to offer detailed, constructive feedback to help improve it.  

\vspace{0.5cm}
\textbf{Here is the email for review:}  
\{email\}
\end{tcolorbox}
\caption{Prompt for expert-level email review and constructive feedback.}
\label{fig:email-review}
\end{figure*}

\begin{figure*}[ht]
\centering
\scriptsize
\begin{tcolorbox}[colback=blue!5!white,colframe=blue!75!black,title=Email Revision]
You are an expert at revising emails. You will be provided with:  
1. An original email.  
2. A set of suggestions on how to improve that email.

\vspace{0.5cm}
\textbf{Objective:}
\begin{itemize}
    \item Transform the original email into a new version that incorporates the given suggestions.
    \item Ensure the final output strictly follows the JSON structure below.
\end{itemize}

\vspace{0.5cm}
\textbf{Output Format:}  
Your response must be a JSON object with these keys:
\begin{itemize}
    \item \texttt{sender\_name}: The name of the sender.
    \item \texttt{from\_address}: The sender's email address.
    \item \texttt{to\_address}: The receiver's email address.
    \item \texttt{send\_time}: The time the email is sent in RFC3339 format without a time zone.
    \item \texttt{subject}: A concise and relevant subject line.
    \item \texttt{body}: The complete email body text.
\end{itemize}

\vspace{0.5cm}
\textbf{Instructions:}
\begin{itemize}
    \item Retain any key information from the original email.
    \item Incorporate the suggestions provided where relevant.
    \item The final email body should reflect a polished, improved version of the original.
    \item Do not add any additional keys; only use the five specified keys.
\end{itemize}

\vspace{0.5cm}
\textbf{Original Email:}  
\{original\_email\}

\vspace{0.3cm}
\textbf{Suggestions:}  
\{suggestions\}
\end{tcolorbox}
\caption{Prompt for revising and improving emails based on specific feedback.}
\label{fig:email-revision}
\end{figure*}

\section{Details of Extrinsic Evaluation}

\subsection{Tasks}
\label{sec:tasks}

\textbf{Email Categorization.} Classify an email into one of eight predefined categories: Professional, Academic, Personal, Promotional, Financial, Social, Spam, or Shopping. Labels for public datasets are obtained via majority voting from three independent GPT-4o API calls, while labels for private datasets are generated using our internal classifier. Accuracy and macro F1 scores are reported.

\textbf{Email Drafting.} Given the email’s subject, sender, and receiver, generate the body of the email. The generated content is compared against the ground-truth email using ROUGE-L \citep{lin2004rouge} and BERTScore \citep{zhang2019bertscore}. Note that some datasets do not contain subject lines for emails; such datasets are excluded from this task.

\textbf{Question Answering.} Given a complete email and a factual question about its content, generate an answer based solely on the email. Questions and reference answers are generated via GPT-4o for public datasets and our internal model for private datasets. An additional LLM-based verification step is used to validate the correctness of generated answers. Performance is measured using ROUGE-L and BERTScore.

\textbf{Next Message Prediction.} This task includes two settings: (1) generation, where the goal is to produce the next message in a conversation thread, and (2) classification, where the model selects the correct next message from a set of ten candidates. In the classification setting, the distractor candidates are sampled from the 100 messages in the training set that are most similar (in embedding space) to the correct next message. Cosine similarity is used to measure closeness in embedding space. We use BERTScore for the generation setting and accuracy for the classification setting.

\subsection{Test Datasets}
\label{sec:test-datasets}

\textbf{Enron} \citep{klimt2004enron} comprises approximately 500,000 emails from around 150 Enron employees, primarily senior executives, collected during the company's collapse in 2001. 10,000 emails are randomly selected as the test set.

\textbf{Human-Gen Phishing} \citep{greco2024llmgen} consists of 2,000 most recent emails from Nazario and Nigerian Fraud datasets \citep{al2024phishing}.

\textbf{Private} and \textbf{Private w/o Spam} refer to our proprietary datasets comprising emails and text messages. To construct the latter, we applied an internal spam classification model to filter out spam content, yielding a subset containing only non-spam messages. It is a faaithful reflection of digital footprints.

\textbf{W3C-Emails} \citep{zhang2021emailsum} comprises email communications from the World Wide Web Consortium's (W3C) public mailing lists. Collected through a crawl of W3C's public sites in June 2004, the dataset includes approximately 174,000 emails, of which 10,000 are used as the test set.

\textbf{DailyDialog} \citep{li-etal-2017-dailydialog} is a human-written, multi-turn dialogue dataset that captures daily communication across topics such as relationships, work, and health. For our evaluation, we use its test set, which consists of 1,000 conversations.

\textbf{Persona-Chat} \citep{zhang2018personachat} comprises over 10,000 dialogues totaling more than 160,000 utterances. Conversations were crowd-sourced via Amazon Mechanical Turk, where participants were instructed to embody their assigned personas during the dialogue. 10,000 conversations are selected in our evaluation.

\textbf{u-sticker} \citep{chee2025usticker} comprises approximately 370,200 sticker instances, with 104,000 unique stickers, collected from 22,600 users across diverse conversational contexts. It was gathered from various online chatting platforms. A subset of size 10,000 is used in this work.

\subsection{Implementation Details}
\label{sec:implementation-details}
For each task, we fine-tune the Mistral-7B-v0.1 \citep{jiang2023mistral7b} model on a given synthetic dataset and evaluate its performance on real-world datasets. To enable efficient fine-tuning, we employ Low-Rank Adaptation \citep{hu2022lora} with configuration parameters \( r = 8 \), \( \alpha = 16 \), and dropout = 0.05. The learning rate is $5 \times 10^{-5}$ with a linear scheduler. To ensure a fair comparison across datasets, we uniformly sample 4,000 training examples and 1,000 validation examples from each synthetic dataset. Models are fine-tuned for two epochs, and the checkpoint with the lowest validation loss is selected for final evaluation.

\section{Cost Analysis}

Here we briefly estimate the cost for creating the dataset. For at most 5 generator–critic cycles, using current LLM API pricing (e.g., Gemini 1.5 Pro: 2.5 USD per 1M input tokens, 10 USD per 1M output tokens), each generation of ~1,500 input + 1,500 output tokens costs about \$0.019. With 2 generations per round and up to 5 rounds, the upper bound per artifact generator agent is ~0.19 USD, and considering event and persona agents, the upper bound of the end-to-end cost is about 0.57 USD per artifact.

\end{document}